\documentclass[conference]{IEEEtran}
\IEEEoverridecommandlockouts
\usepackage{cite}
\usepackage{amsmath,amssymb,amsfonts}
\usepackage{algorithmic}
\usepackage{graphicx}
\usepackage{tabularx}
\usepackage{textcomp}
\usepackage{xcolor}
\def\BibTeX{{\rm B\kern-.05em{\sc i\kern-.025em b}\kern-.08em
    T\kern-.1667em\lower.7ex\hbox{E}\kern-.125emX}}
    
\usepackage{eso-pic}
\newcommand\AtPageUpperMyright[1]{\AtPageUpperLeft{
 \put(\LenToUnit{0.5\paperwidth},\LenToUnit{-1cm}){
     \parbox{0.5\textwidth}{\raggedleft\fontsize{9}{11}\selectfont #1}}
 }}
\newcommand{\conf}[1]{
\AddToShipoutPictureBG*{
\AtPageUpperMyright{#1}
}
}
    
\begin{document}

\title{Towards Conversational Humour Analysis and Design}

\author{\IEEEauthorblockN{
Tanishq Chaudhary*,
Mayank Goel*,
\thanks{* These two authors contributed equally}
Radhika Mamidi
}
\IEEEauthorblockA{
LTRC, International Institute of Information Technology, Hyderabad\\
Email:
\{tanishq.chaudhary, mayank.goel\}@research.iiit.ac.in, radhika.mamidi@iiit.ac.in,
}}

\maketitle
\conf{This work was presented at the 11th Humor Research Conference, 26-27 February 2021}

\begin{abstract}
Well-defined jokes can be divided neatly into a setup and a punchline. While most works on humor today talk about a joke as a whole, the idea of generating punchlines to a setup has applications in conversational humor, where funny remarks usually occur with a non-funny context. Thus, this paper is based around two core concepts: Classification and the Generation of a punchline from a particular setup based on the Incongruity Theory. We first implement a feature-based machine learning model to classify humor. For humor generation, we use a neural model, and then merge the classical rule-based approaches with the neural approach to create a hybrid model. The idea behind being: combining insights gained from other tasks with the setup-punchline model and thus applying it to existing text generation approaches. We then use and compare our model with human written jokes with the help of human evaluators in a double-blind study. 

\end{abstract}


\section{Introduction}
While there is no single theory of humour that can explain what makes jokes funny, or what makes a joke, a joke; we have multiple prominent theories that explain the core concepts of humour which have laid way to work such as ours. The core theory on which modern computational humour is currently building on is called the Incongruity theory, the idea that for a joke to be a joke, it needs to have compatible but contrastive “scripts”, or ideas. Primarily focusing on the object of humor, this theory sees humor as a response to an incongruity, a term broadly used to include ambiguity, logical impossibility, irrelevance, and inappropriateness. Take the following joke as an example.
\\
\\
\emph{``Two fish are in a tank. One of them asks the other, “Do you know how to drive this thing?”"} \cite{ritchie}
\\
\\
In this joke, we see the two opposing ideas, the idea of an implied fish-tank, and that of the absurdity of a vehicular tank. Thus, an idea based on Incongruity theory can be formalized in two scripts, for short jokes: a setup, and a punchline. This form of formalization is familiar in context of two-liners, or even in Stand-up comedy, where there is an absurd question, or a situation, followed by a perfectly compatible but unexpected answer. From a semantic perspective, Raskin’s script-based semantic theory of humour is a formal theory which also gives insights into what makes a joke, a joke. For it, Raskin states two conditions: (i) The text is compatible, fully or in part, with two different scripts (ii) The two scripts with which the text is compatible are said to overlap fully or in part on this text. \cite{raskin}

These theories, while formal, cannot be applied directly, as specific examples may be useful for understanding a joke, but the pragmatics and real-world knowledge to understand texts is yet not possible by a computer. While existing methods for text generation are useful, utilizing such theories for specific humor generation may lead to better results.

\section{Dataset}
We used an existing dataset of 200k jokes from r/jokes on kaggle\footnote{https://www.kaggle.com/abhinavmoudgil95/short-jokes}. The dataset was cleaned according to the need of each task, as mentioned under their own headings. The jokes have several quirks which are unique to Reddit, such as use of asterisk for emphasis, and jokes that are directly related to the Reddit community. For classification, the non-humorous data used is content from Firefox discussion forum\footnote{Can be accessed from nltk.corpus’ webtext}.

\section{Humor Classification}

\subsection{Related Work}
Humor Classification is one side of the larger idea of understanding computational humor. To accurately understand the current state of humor on social media, we turned to Reddit. We use a statistical feature based machine learning model. 

Work by Mihalcea et al., \cite{mihalcea} takes a look at one-liners, in comparison to other newspaper headlines and a general corpus. Work by Zhang et al., \cite{zhang} explores humorous tweets in comparison to other non-humorous tweets, whilst exploring numerous features. 

\subsection{Motivation}
Our work focuses more on the current style of humor, as taken from the Reddit style of jokes. This makes the task significantly challenging, since the jokes are neither instances of stock humor or have distinctive features which can be found in tweets only. We instead deal with jokes playing on certain real world knowledge. We explore prominent features pointed out by the previous works in light of the Reddit dataset.

\subsection{Features}
\subsubsection{Morpho-Syntactic Level}
We first look at the Part Of Speech (POS) tags to make the judgement. We use Spacy's POS tagger for English. 
Zhang et al., mentions the use of differences in part of speech ratios for verbs, nouns, pronouns, proper nouns and modifiers. We decided to test all of the above, in the context of our dataset.

\subsubsection{Lexico-Semantic Level}
Mihalcea et al., mentions the large amount of usage of adult slang in humorous texts. The Reddit taste of such kind of humor is a bit different and we could not rely on any external tools. We decided to create a list of slang words on our own. There are also hidden instances of slang. The following example has not one, but two such instances.
\\
\\
\emph{
``my sister can't stop having sex. ... i think she's addickted."
}
\\
\\
We predict this will be the largest and most weighted indicator of humor content. We also look at antonyms pairs. This idea directly draws from the incongruity theory mentioned earlier, and has been explored by Mihalcea et al. The following is an example of such antonyms.
\\
\\
\emph{
``smaller babies may be delivered by a stork but the bigger heavier ones are delivered by a crane"
}
\\

\subsubsection{Pragmatic Level}
Zhang et al., mentions the use of discourse connectives. Many of the Reddit jokes are not one-liners or even limited to 140 characters (as for tweets), since there is no such restrictive limit in Reddit. Infact, due to the presence of longer jokes, we would expect a greater use of discourse connectives, to have a more `proper' punchline at the end. We use a hand-crafted list of popular discourse markers. The following is an example of a long joke using a single connective.
\\
\\
\emph{
``I'd like to think that my girlfriend and I have a relationship that is above being forced to buy simple gifts as part of a made up holiday that exploits working class people through the commercialism of enormous corporations ... But I'd also like to get laid tomorrow night, so Walgreens after work it is."
}
\\
\subsubsection{Affective Level}
The idea behind using affective features is that humor evokes a certain positive or negative emotion omnipresent. We use SenticNet version 6.0 \cite{sentic} to take into account the polarity of the sentence.
\\
\\
\emph{
``Four score and seven years ago our fathers brought forth on this continent, a new nation, conceived in Liberty, and dedicated to the proposition that all men are created equal. ... D J Trump: Fake News!"
}
\\
\\
Are some examples of negative and positive polarities.

\subsection{Results}
As one can see from the images (present at the end), each of the features shows very little difference from the general versus the humorous corpus. We see slangs as a very good differentiator. Using sklearn \cite{sklearn}, we use all the features mentioned and train using various machine learning models.

\begin{table}[]
\caption{Humor Classification: Model comparisions}
\centering
\begin{tabular}{|l|l|l|}
\hline
\textbf{Model} & \textbf{F1} & \textbf{Accuracy} \\
\hline
Logistic Regression & 0.58 & 0.605 \\
\hline
Gaussian Naive Bayes & 0.59 & \textbf{0.625} \\
\hline
Support Vector Classifier & \textbf{0.63} & 0.62
\\\hline
\end{tabular}
\end{table}

\subsection{Conclusion and Analysis}
We see that while each feature contributes little to the classification problem, these features in combination are able to give us decent results. Of the models, Support Vector Classifier outperforms other models, since it is able to capture the interactions between the features, unlike other models. At the same time, we must take into account that humor on Reddit also relies a lot on real world knowledge, for which a simple feature based machine learning model is rather insufficient.

\section{Humor Generation}

\subsection{Related Work}
Generating Humour is a very difficult task, and modern applications work as search engines on human-vetted jokes for generating humour. While there has been some progress into generating puns, pun generation is a morpho-syntax task, and generating discourse that resembles jokes (a task that is difficult even for humans), is a task that needs a combination of various approaches to be done successfully.

In Winters et al.\cite{winters}, they have a defined methodology for a general method for humor generation. It used semantic distances between tokens for generation, which is a novel idea that can be abstracted into using masked language models. In Strategies for Structuring Story Generation by Fan et al.,  \cite{fan}, we see the idea of substituting words in text generation. Petrovic \cite{petrovic} uses unsupervised learning for text generation. Looking at this form of research, we wanted to try out a classic neural model and then work forward on solving challenges that come up. At the time of starting this project, our expectation was to get coherent results in the neural stage itself.

\subsection{Baseline: Markov Chains}

\subsubsection{Motivation}
Language can be seen as a sequence of tokens. By using stochastic modelling, we want to create a model that takes in a seed sequence of words - which we call the setup - and then generate entire jokes from it. Discrete time Markov chains allow us to create that very structure: taking in an input sequence of words and then figuring what is the next best word; and then doing this till an ending token is reached. This relies on the simple idea that words which appear together in one particular context, are more probable to appear in other similar contexts. 

\subsubsection{Language Modelling}
We combine the power of probabilistic stochastic models with n-grams. Where grams mean tokens; which could either be word or character level. This gives us an idea of the transition probabilities, for each such sequence of grams. More specifically, we consider the $(n-1)^{th}$ order Markov assumption. It allows us to consider the previous $n-1$ states only, instead of the entire preceding sequence. We look at both the word and the character level for the generation.

Markov assumption for n-gram:

\begin{align}
    \nonumber
    Pr(X_i | X_{i-1}, X_{i-2}, X_{i-3}, ... , X_{1})
    \\\nonumber
    \approx Pr(X_i | X_{i-1}, X_{i-2}, X_{i-3}, ... , X_{i-n+1})
\end{align}

Here, $X_i$ is the $i^{th}$ token. Thus, instead of going from the first token to the current one, we can simply take the last $n-1$ tokens.

\subsubsection{Results}

\begin{table*}[]
\centering
\caption{Humor Generation: Markov Chains}
\begin{tabular}{|l|l|l|}
\hline
\textbf{Level} & \textbf{Gram} & \textbf{Sample} 
\\\hline
Character & 2 & \begin{tabular}[c]{@{}l@{}}A.\\ I.\end{tabular}
\\\hline
Character & 3 & Why daisinguids of my partionald there you could him. 
\\\hline
Character & 4 & \begin{tabular}[c]{@{}l@{}}What didn't like answer culture.
\\
Knock Knights for put these doctors on to probably.\end{tabular} \\
Character & 5 & Why couldn't under 10 million date? . \\\hline
Character & 6 & \begin{tabular}[c]{@{}l@{}}What do you know that shit got two gay man is a duck say to ruin a joke a lightbulb?
\\
Did you hear about the floor of things Dickens wear glass clown.
\\ 
Went to say really *stepped on the explodes white.\end{tabular} 
\\\hline
Character & 12 & What did the physicists so bad at soccer? . 
\\\hline
Character & 20 & \begin{tabular}[c]{@{}l@{}}What I'm doing to occupy my free time now that the majority of car accidents happen in Russia.
\\
Thank you student loans, for helping me get through college ... I don't think the speed was why I was arrested though…
\\
What's the difference between Rosh Hashanah and Yom Teruah?"  \\
And he said to me, he said:  \\ "Oh, about fifty bucks."...\end{tabular} \\\hline
Word & 2 & A friend of mine asked if I should use to cut by . \\\hline
Word & 3 & \textbf{Teacher : All Idiots Stand Up A boy stand up Teacher : so are you familiar with my parodies ? Yeah my PAIR O DEEZ NUTS ! ! }
\\\hline 
Word & 4 & \begin{tabular}[c]{@{}l@{}}Why are there so many trees along the Champs - Élysées ? ... Because ze Germans like to march in ze shade .\\ What do you call a midget with Down Syndrome? ... you call him a little slow\end{tabular}

\\\hline

\end{tabular}
\end{table*}

As can be seen from table II, when considering character based models we see a steady improvement in the quality of the text generated. Going from $n = 2$ to $n = 6$, it learns the correct spellings and starts to figure grammar for short sequences, although still making little to no logical sense. We also see the case of \emph{ ``Went to say really *stepped on the explodes white.”} which is interesting to note as it encapsulated the `*’ symbol used commonly, for expressing emotions or sound effects in text. However, due to the small range of its sight, it has failed to close the `*’ symbol, on the other side of the word. This is fixed as we proceed towards $n = 20$.

However, as we go beyond, we see no significant improvement in the logical sense of the sentence. Clearly, a character based model is rather insufficient for text generation, let alone for humor generation.

When it comes to word level Markov models, we see two order sentences are insufficient, but order 3 sentences give us the best results from Markov chains. The joke, \emph{``Teacher: All Idiots Stand Up. A boy stand up. Teacher: so are you familiar with my parodies? Yeah my PAIR O DEEZ NUTS!!”} is novel and is created with the merging of the two parent jokes:
\\\\
\emph{``Teacher: All Idiots Stand Up A boy stand up Teacher: so are you an idiot ? Boy: No I can't bear you standing alone madam... ... "}
\\\\
\emph{``What is Weird Al Yankovic's favorite pick up line? ... Hey, so are you familiar with my parodies? Yeah my PAIR O DEEZ NUTS!!!",
}
\\\\
Which were not as funny. If we go to $n \ge 4$, we see that the model simply starts to memorize word sequences. 

\section{Humor Generation: Word Level LSTM}

\subsection{Motivation}
We clearly saw that a big component missing from the statistical language modelling is logic, and the generation of coherent sentences. It may be due to the sparsity of the data present; there simply may not be enough contexts to handle different setups. Instead of manually designing backoffs and other smoothing formulae, we decided to use neural techniques \cite{potash}. They allow for sparsity of the data, since they scale easily to larger vocabulary sizes: which could mean allowing for more obscure contexts.

\subsection{Neural Language Modelling}
LSTMs (Long Short Term Memory) are a particular type of RNN (Recurrent Neural Networks) \cite{sutskever} which are used for sequential data, with the added bonus of them having the ability to forget. This allows us to bypass the computational errors of vanishing and exploding gradients when training on longer sequences of data.

\subsection{Experiments}
\subsubsection{Data Preprocessing}
We use Keras’ regex tokenizer, since for such a large dataset, using any other library would take lots of computational power. Using regex, we also ignore punctuations that make the model unstable to deal with (explained later).
\subsubsection{Model Architecture}
Since the dataset is large enough in size, we train our own embeddings. We then have 2 LSTM layers, with the hope that the first captures the basic syntax level meaning and the second layer captures the deeper semantics. We also introduce dropout, in an effort to prevent overfitting of the model. We then follow it by two fully connected layers, which select what word should come after a given sequence.
\subsubsection{Results}
We observe that for sequences of length less than 4, the model would underfit. If we went greater than or equal to 4, the model would overfit. Most of the jokes generated are thus not novel. We also observe that the jokes degenerate after a certain length of text is generated. The model starts to produce repeating sequences, type of which can vary depending on the presently generated sequence.

\begin{table*}[t]
\centering
\caption{Humor Generation: Comparing LSTM jokes with their nearest parent}
\begin{tabular}{|l|l|}
\hline
\textbf{Closest Original Joke} & \textbf{LSTM Generated Joke (truncated to 100 characters)} \\
\hline
what's got 90 balls and screws old women? ... bingo! & what's got 90 balls and screws old women bingo of the world of the world of the world of the world \\\hline
"limericks eh ? ... there was this girl from boston, mass. & limericks eh there was this girl from boston mass she wade into the sea and wet her ankles it doesn' \\\hline
i found waldo ... he's mexican now. & i found waldo he's mexican now i don't know how much i can tell you a joke about a black guy\\\hline
\end{tabular}
\end{table*}

\subsection{Challenges Faced}
We saw that for lower sequence lengths, the model was extremely unstable, and would often degenerate to using articles only. If we left the punctuations in, the model would simply start producing a sequence of ‘.’ s. This is why we had to use a regex tokenizer that removed all such punctuations, forcing the model to learn. We also observed that the model was very prone to overfitting. Even after rigorous experimentation using many hyperparameter configurations, we did not find any decent results. The jokes it has generated are very similar to the ones seen in the dataset, with very few novel creations. In other cases, the grammar was barely in place, although it got the basic phrases right. 

\section{Humor Generation: Hybrid Method}
\subsection{Motivation}
A major challenge of neural methods is the difficulty in maintaining coherent grammar. On the other hand, using templates for joke generation is a staple technique in Humor Generation. This is aided by the intrinsic behavior of humans to detect patterns and meaning. We thus built a model that extracts templates, and then fills in the templates using BERT, and a model particularly suited for the task since it is a Masked Language model. 

\subsection{Template Extraction}
For template extraction, the immediate challenge is trying to understand which words are constituent to a joke’s meaning, and in which order. Removing all content words from a sentence would generate templates, however the essence of the joke will be lost and so is the coherency. The idea is to take in two factors: How important is that entity to the joke? The lesser the importance, the earlier it will get replaced. How frequent is the word in natural language? The less frequent, the more “unique” it is, thus more replaceable.
\\
\\
Consider:\\
``Why did the X cross the road?”\\
``Why did the chicken cross the X?”\\
``Why did the chicken X the road?”\\
\\
Here, the first sentence changes the least meaning, allowing multiple entities to be filled in as part of the punchline. The second sentence is less flexible than the first, as arguably “road” is more essential to the joke than “chicken”. The third sentence having the verb can be seen to encode the most important semantic information, of crossing the road. Thus, we looked at the dependency graphs of several such jokes, and came up with weights of our own that should approximately hold. 

As part of this, we removed all Adjectives directly associated with a noun, and the weights for Adjectives is only for non-immediate entities like “Life is good”. 

\begin{table}[]
\caption{Weights for dependency tags}
\centering
\begin{tabular}{|l|l|}
\hline
\textbf{Weight} & \textbf{Dependency tag} \\
\hline
10 & Named Entities \\
\hline
5 & Subject \\
\hline
4 & Indirect Object \\
\hline
3 & Direct Object \\
\hline
2 & Adjective \\
\hline
1 & Verb \\
\hline
\end{tabular}
\end{table}

The second part of calculating scores is the word frequency in natural language. The less common the word, the less important it is to the joke. We used a dataset of 10k words\footnote{https://github.com/first20hours/google-10000-english/} ranked in terms of word frequency, and calculated the final weight by:
$$score_{total} = score_{grammar}\times log_{10}(score_{wordfrequency})\times 2.5$$

The log is taken of the word frequency score to scale it properly, as without it the natural frequency will outweigh the grammar score.

\subsection{Infilling}
The major need for any model that can do infilling is that it needs to be able to consider nearby contexts. Thus, the choice of using BERT \cite{bert}, a bi-directional masked language model, which uses transformers and is context-dependent. Thus, a template will look like:

\emph{``Why did the [MASK] cross the road?”} and utilizing BERT, we can print out tokens for infilling.

A major advantage in using a transformer for this task, especially with pre-trained models is the capacity to fine-tune. Thus, applications can be fine-tuned using corpora from different domains, and lead to varied forms of jokes. 

\subsection{Results}
\begin{table}[htb]
\centering
\caption{Humor Generation: BERT jokes}
\begin{tabular}{| p{0.2 \linewidth} | p{0.2 \linewidth} | p{0.2 \linewidth} | p{0.2 \linewidth} |}
\hline
Original Joke & Template & BERT Joke & Comments \\
\hline
What s long and hard and full of semen ? A Submarine . & what s long and {[}MASK{]} and full of {[}MASK{]} ? a {[}MASK{]} . & what s long and thin and full of life ? a smile . & Novel joke \\
\hline
How do Mexicans feel about Trumps wall ? They 're already over it . & how do {[}MASK{]} feel about {[}MASK{]} wall ? they {[}MASK{]} already over it . & how do you feel about the wall ? they were already over it . & Minimal change in semantic meaning, nouns move to pronouns \\
\hline
What 's the seamonster 's favourite meal ? Fish \& Ships & what 's the seamonster 's {[}MASK{]} ? fish {[}MASK{]} ships & what ' s the seamonster ' s name ? fish and ships & Improper template extraction\\
\hline
\end{tabular}
\end{table}

This sort of infilling is done by utilizing the [MASK] tokens, and looking at all other words at once. As we see here, by preserving the original template, we maintain very coherent grammar, and by infilling, we gain legible semantic meaning.

\section{Evaluations}

We had 5 human evaluators, each classifying 50 jokes each. The evaluation script would randomly select jokes either from the Reddit dataset or the hybrid model generated jokes dataset.

On basis of the Confusion Matrix, we can see that the recall (83.82\%) is lesser than the precision (91.93\%) , which shows that the human jokes themselves aren't always funny and coherent, which can be expected. At this point, the precision isn't amazing, however, considering the fact that the jokes being created are novel, it is a step towards multi-domain joke generation, where various forms of humor can be fine-tuned, and shows promise for joke generation using template infilling. 

\begin{table}[htb]
\centering
\caption{Confusion Matrix for manual evaluation of jokes}
\begin{tabular}{|l|l|l|}
\hline
\textbf{Confusion Matrix} & \textbf{Predicted Computer} & \textbf{Predicted Human} \\\hline
Actual - Computer & 114 & 10 \\\hline
Actual - Human & 18 & 108\\\hline
\end{tabular}
\end{table}

\section{Conclusions and Future Work}
\textbf{Humor Classification}: We note that the features were put at all different levels, of any form of analysis. We went from exploring phonetic features, all the way up to discourse markers. It is evident that the results are not as good as compared to accuracies of one-liners from Mihalcea et al., or twitter from Zhang et al. We can only reach around 63\% accuracy. On the other hand, this is indicative of the fact that the features do contribute to the classification task. Be it pulling out antonym pairs or usage of slangs; everything contributes to a better classification of humorous versus non-humorous texts. For future work, one can work on more sophisticated features that are able to capture the intricacies of humor. One can also look at neural approaches for better results.

\textbf{Humor Generation: Markov Chains}. We saw how the use of small n is not able to form grammatically correct sentences (or even words with correct spellings for character level). And, if we increase it too much, we see how the model starts to repeat the jokes and overfit. We found word level models with order 3 as the best configuration. We can thus conclude that word based models are significantly better character based models, and can even touch the levels of “jokness”, although on an average still gravely restricted in the logical sensibility.

\textbf{Humor Generation: LSTM Model}. The model was unable to generate grammatically correct sentences, though was able to pick up correct phrases. LSTMs are supposed to be good at long range sequences, and we see that that is not the case when it comes to generation with limited data, especially when one talks about humor generation. Thus, the model is woefully inept to generate text, let alone jokes. 

\textbf{Humor Generation: Hybrid Model}. The model benefits from the template method by having a coherent grammar, and also from the powerful BERT model to give it coherency. The biggest advantage of this method is the possibility of fine-tuning, where multi-domain contexts can be used to fill in templates, allowing for the possibility of humor. The other challenges are movement from nouns to pronouns, and numerous rules which can be improved upon for template extraction. 

\section*{Acknowledgment}
We would like to thank Saujas Vaduguru and Siddharth Bhat for being available to resolve our doubts. We would also like to acknowledge our human evaluators: Shivansh Subramanian, Sarthak Agarwal, Murali Bhat, Kunwar Grover, Alapan Chaudhari. Our final thanks is to our parents, for their support throughout.

\clearpage

\begin{figure*}
\includegraphics[scale=0.4]{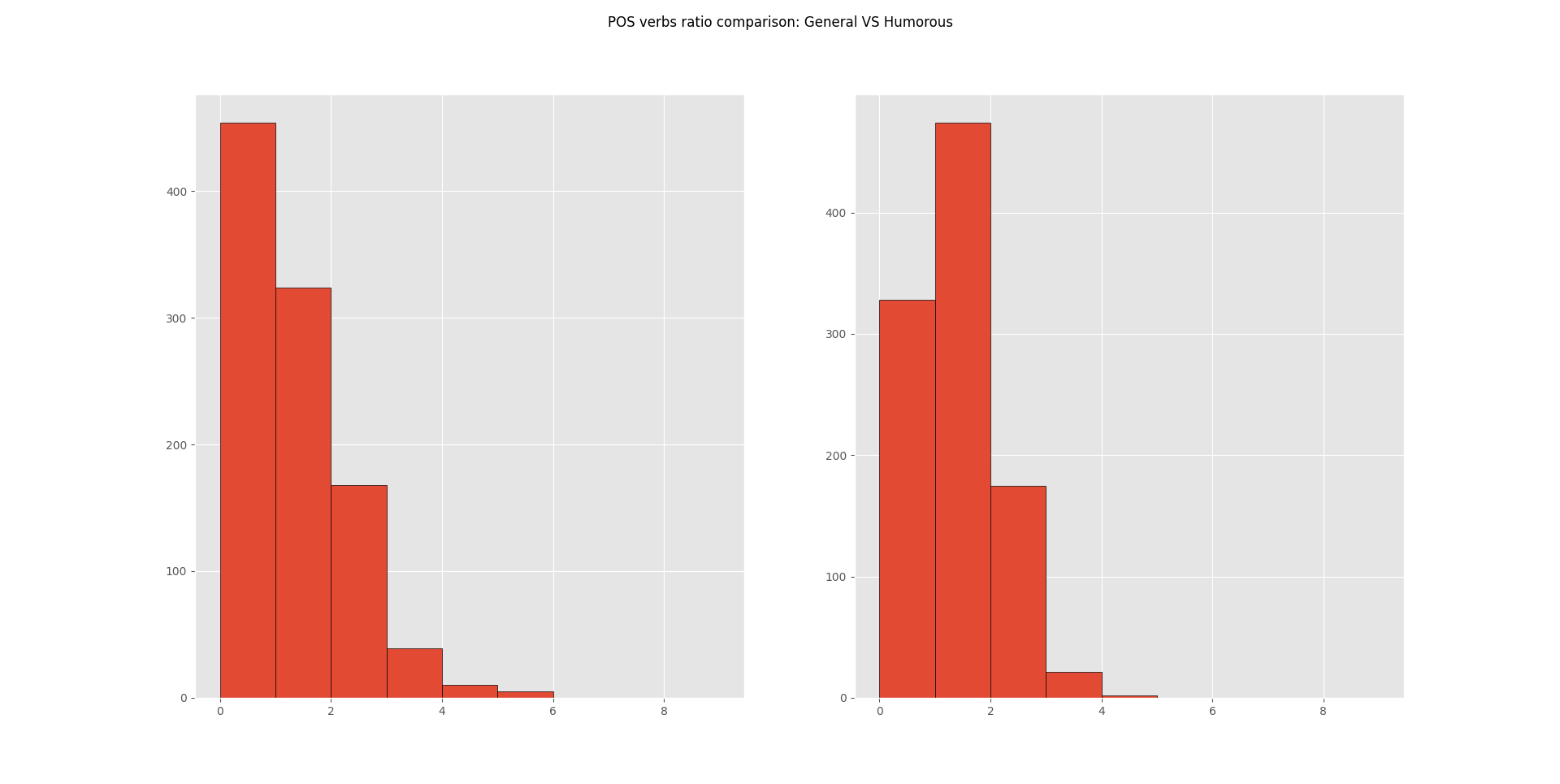}
\end{figure*}

\begin{figure*}
\includegraphics[scale=0.4]{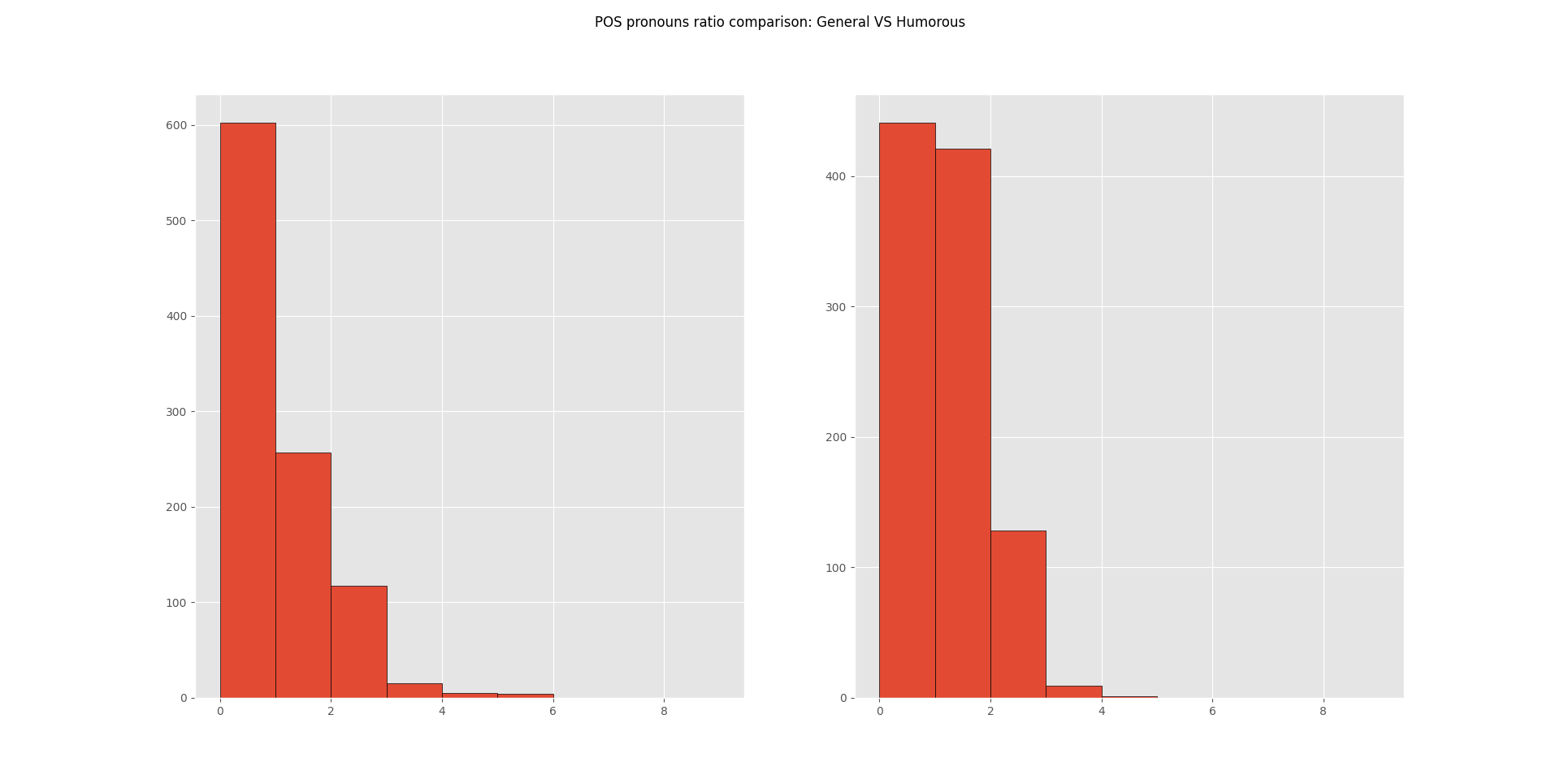}
\end{figure*}

\begin{figure*}
\includegraphics[scale=0.4]{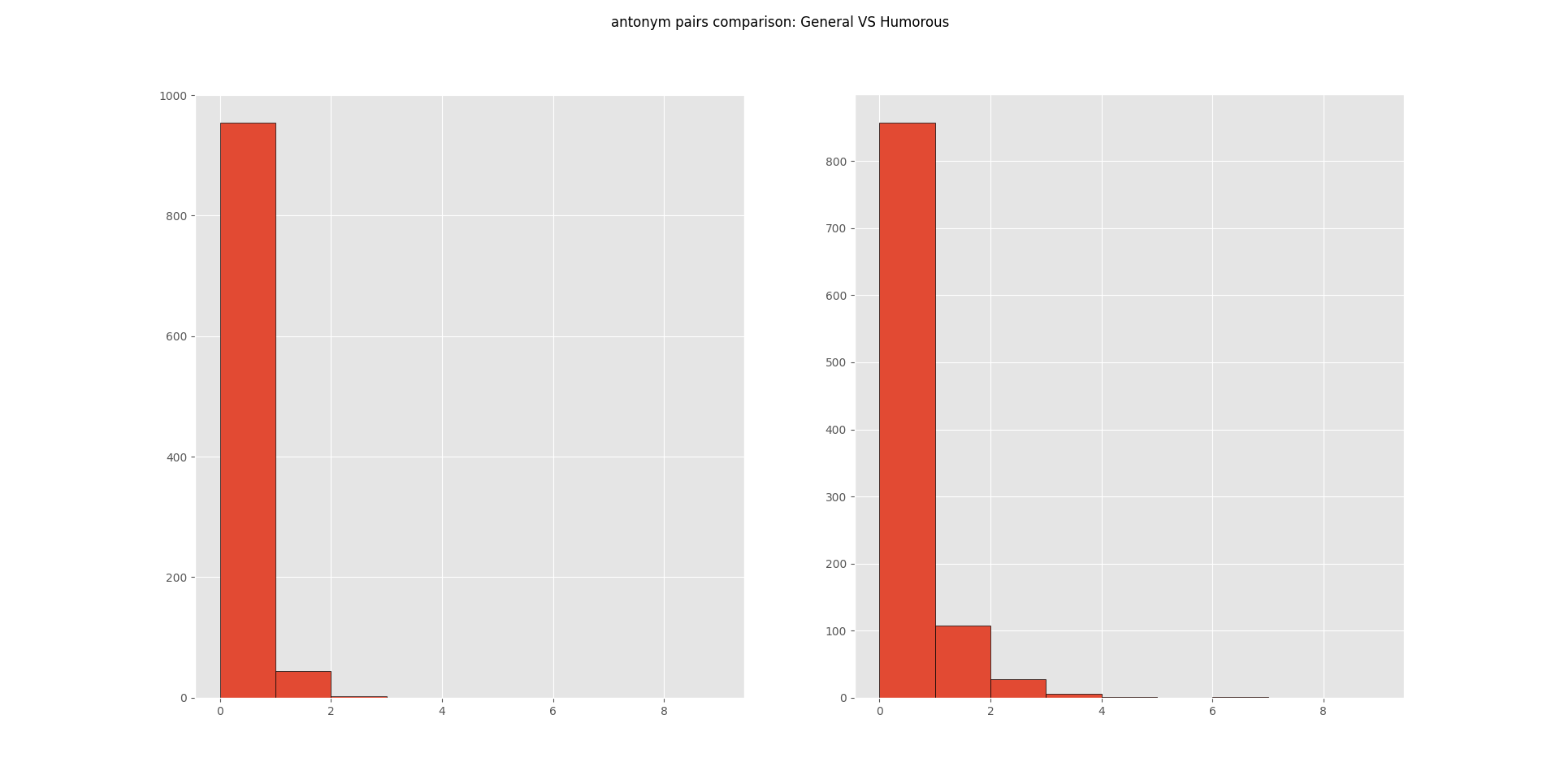}
\end{figure*}

\begin{figure*}
\includegraphics[scale=0.4]{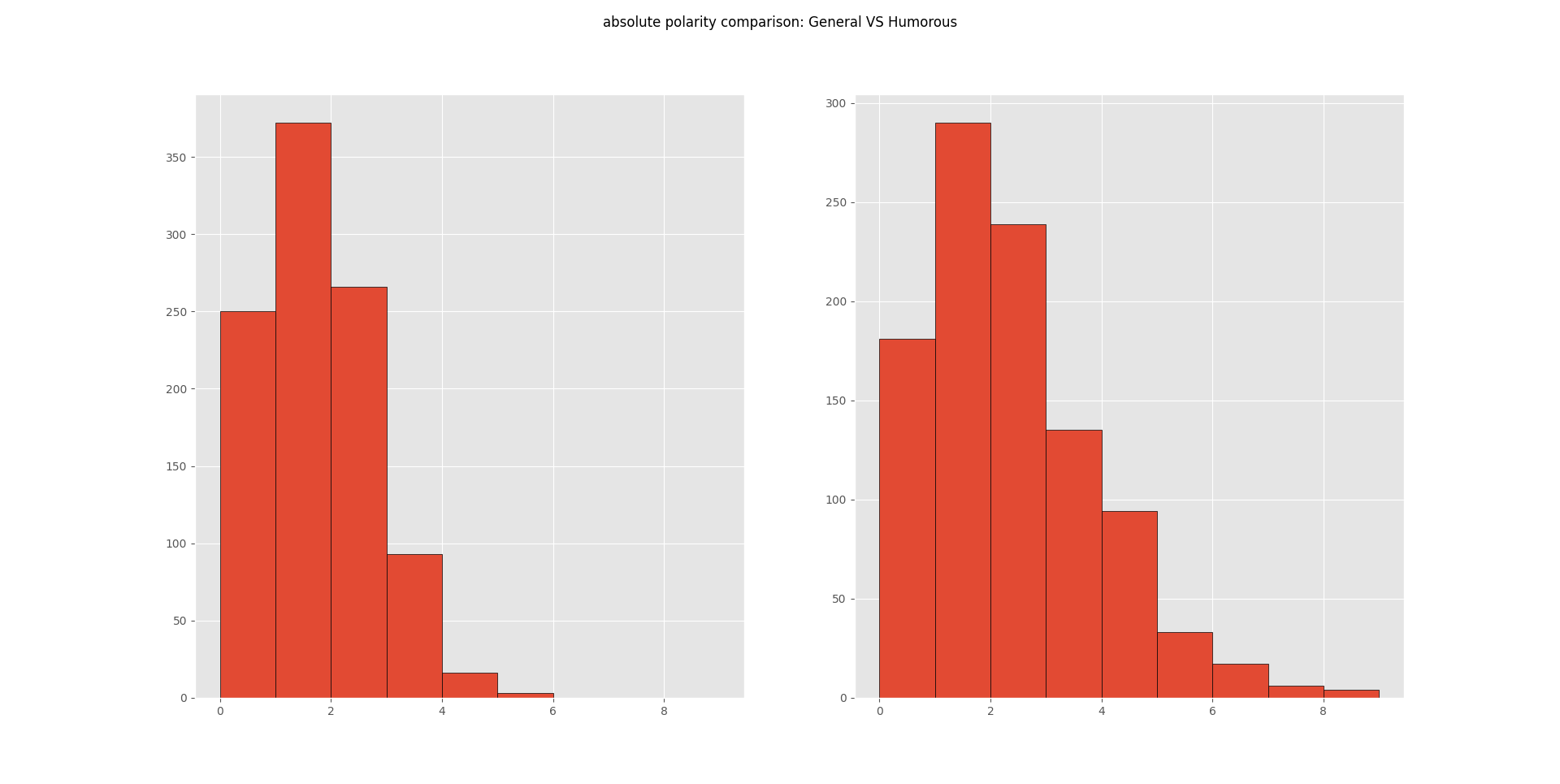}
\end{figure*}

\begin{figure*}
\includegraphics[scale=0.4]{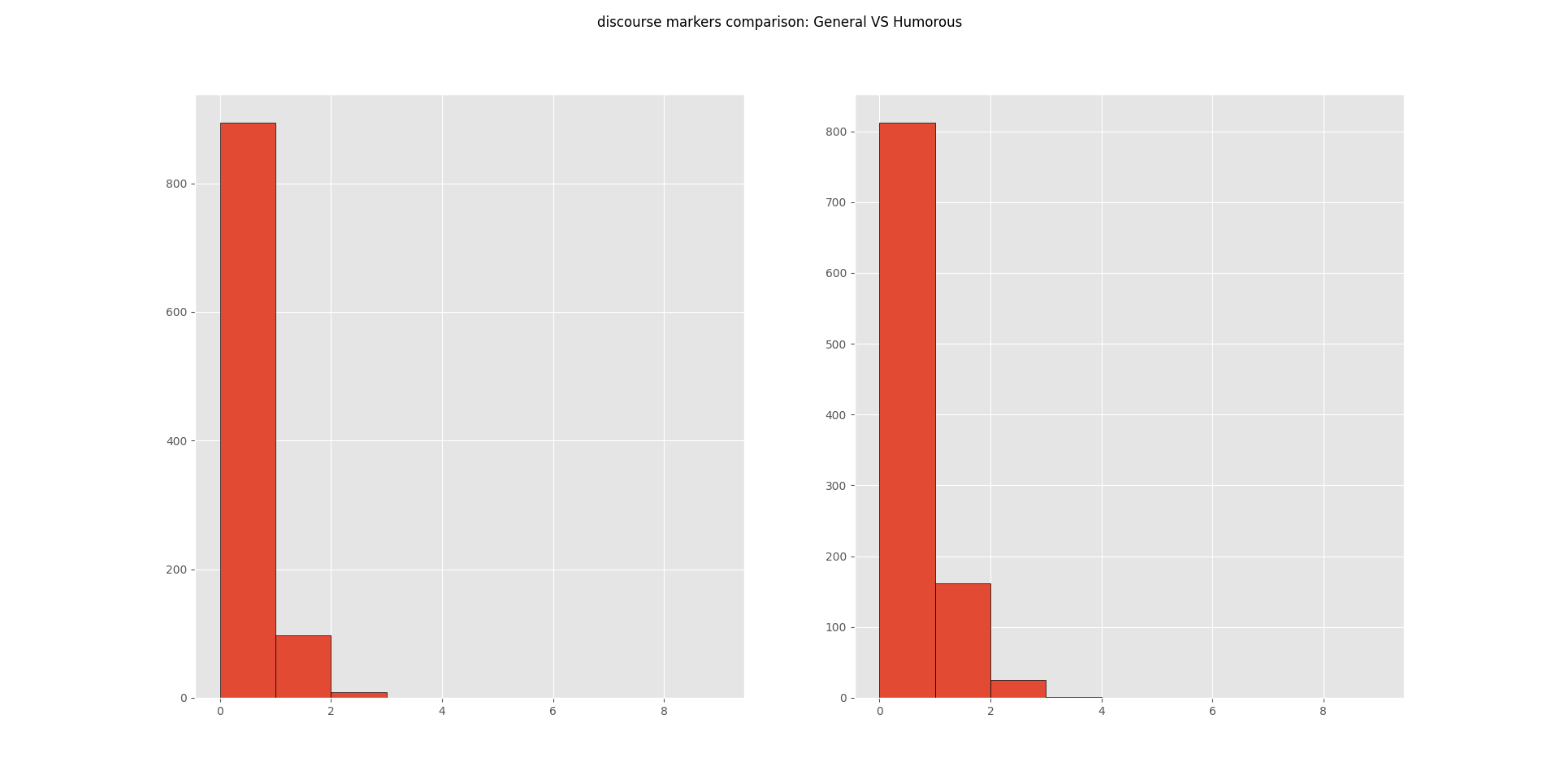}
\end{figure*}

\begin{figure*}
\includegraphics[scale=0.4]{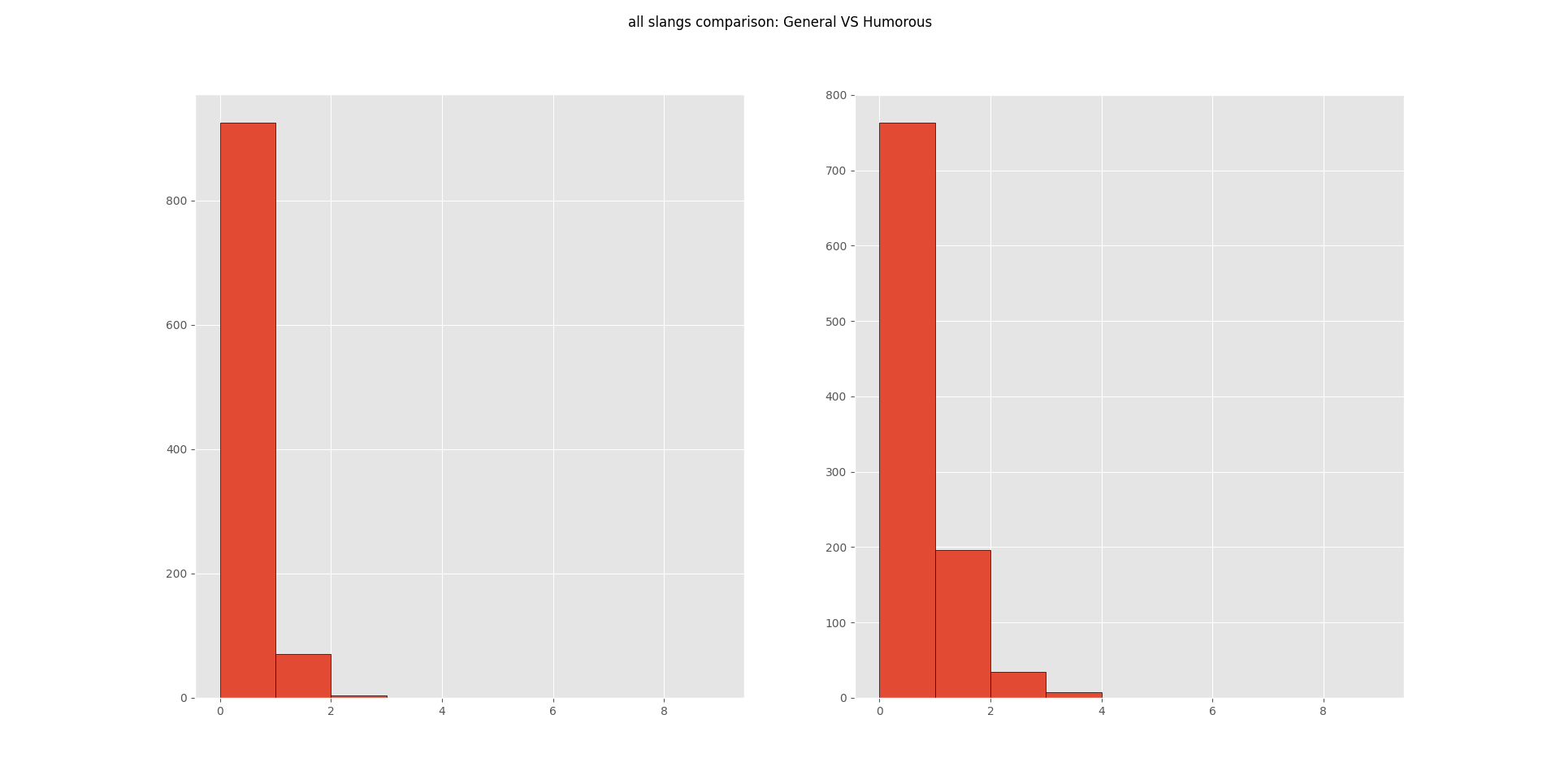}
\end{figure*}

\end{document}